\documentclass[11pt]{article}
\usepackage[final]{acl}

\usepackage[T1]{fontenc}
\usepackage[utf8]{inputenc}
\usepackage{microtype}
\usepackage{inconsolata}
\usepackage{times}
\usepackage{latexsym}
\usepackage{booktabs}
\usepackage{amsmath}
\usepackage{array}
\usepackage{graphicx}
\usepackage{xcolor}
\usepackage{tikz}
\usetikzlibrary{arrows.meta,positioning}
\usepackage{tabularx}
\usepackage{tcolorbox}
\usepackage{xspace}
\usepackage{listings}

\newcommand{\anonrepo}{\emph{anonymized for review}\xspace}
\lstdefinestyle{prompt}{%
  basicstyle=\scriptsize\ttfamily,
  breaklines=true,
  breakindent=1.5em,
  breakatwhitespace=false,
  frame=single,
  framesep=3pt,
  columns=flexible,
  keepspaces=true,
  showstringspaces=false,
  backgroundcolor=\color{gray!7},
  morecomment=[l]{\#\#},
  commentstyle=\itshape\color{gray!70},
}
\renewcommand{\anonrepo}{Code at \url{https://github.com/KRLabsOrg/LettuceDetect}; models and datasets under \url{https://huggingface.co/KRLabsOrg}\xspace}

\title{Beyond Document Grounding: Span-Level Hallucination Detection over Code, Tool Output, and Documents}

\author{
  \textbf{Ádám Kovács\textsuperscript{1}},
  \textbf{Bowei He\textsuperscript{2,3}},
  \textbf{Xue Liu\textsuperscript{2,3}},
  \textbf{István Boros\textsuperscript{1}},
  \textbf{Szilveszter Tóth\textsuperscript{1}},
  \textbf{Gábor Recski\textsuperscript{1,4}}
\\
  \textsuperscript{1}KR Labs,
  \textsuperscript{2}MBZUAI,
  \textsuperscript{3}McGill University,
  \textsuperscript{4}TU Wien
\\
\small{
  \textbf{Correspondence:}
  \href{mailto:kovacs@krlabs.eu}{kovacs@krlabs.eu}
}
}

\begin{document}
\maketitle

\begin{abstract}
Hallucination detection for retrieval-augmented generation (RAG) is usually evaluated on natural-language document evidence. However, grounded generation systems increasingly rely on structured inputs: source code, developer-tool output, markdown documents, tables, and repository metadata. We introduce a unified benchmark for span-level hallucination detection over code, tool output, structured documents, and existing natural-language RAG datasets. The benchmark is built by starting from grounded correct answers, injecting localized hallucinations with exact character labels, and validating the code test split with evidence-based review. Our fine-tuned Qwen3.5-2B detector reaches $0.689$ span-F1 on the unified test set and $0.60$ on the code-agent source, where it substantially outperforms LettuceDetect-large ($0.17$) and the strongest zero-shot LLM judges we evaluated (at most $0.22$). The same model remains competitive on established natural-language benchmarks, with $81.8$ RAGTruth example-F1 and $0.724$ English PsiloQA IoU.
\end{abstract}

\section{Introduction}
\label{sec:intro}
Retrieval-augmented generation (RAG) grounds model outputs in external evidence \citep{lewis2020rag}, but it does not remove the need for verification. A generated answer can still contradict the retrieved context, introduce unsupported information, or cite a reference that is not present in the evidence. Hallucination detection methods therefore ask whether an answer is supported by the supplied context, often at the level of examples, sentences, tokens, or spans \citep{niu2024ragtruth,rykov2025psiloqa,vazquez2025mushroom}.

Most existing benchmarks and detectors study this problem in natural-language RAG, where both the evidence and the answer are usually document text \citep{niu2024ragtruth,belyi2025luna,tang2024minicheck,kovacs2025lettucedetect,rykov2025psiloqa,vazquez2025mushroom}. Real grounded-generation systems are broader: coding agents work over repository files, git history, and test output \citep{jimenez2024swebench}; developer assistants summarize command output and tool observations \citep{kovacs2026squeez}; and research or documentation systems retrieve markdown pages, tables, citations, and structured documents \citep{recski2026aclverbatim,openindex2026wikipedia}. These settings are not well covered by current training data or evaluations: there is no shared span-level benchmark that covers generated code, tool observations, and structured documents under the same verification task.

We study post-generation verification for this structured setting: given an answer that has already been produced, together with its request and context, a detector should flag the parts that the evidence does not support. We frame this at the span level rather than as a whole-answer accept/reject decision, because in code and tool output a single unsupported substring, such as a wrong field, a fabricated method name, a misreported value, or an invented section reference, can change program behavior or mislead a user while leaving the rest of the answer correct. A verifier should therefore point to the unsupported substring, not just reject the answer.

Existing hallucination-detection benchmarks and models leave this setting only partially covered. RAGTruth \citep{niu2024ragtruth}, Luna \citep{belyi2025luna}, MiniCheck \citep{tang2024minicheck}, and LettuceDetect \citep{kovacs2025lettucedetect} verify generated text against retrieved documents. Code hallucination work studies generated snippets \citep{liu2024codehalu,venkatesh2024codemirage}, generation-time divergence \citep{zhang2024collubench}, or agent trajectories \citep{wang2026agenthallu}. These resources are useful, but they do not provide one span-level formulation that covers natural-language RAG, structured documents, source code, and developer-tool output.

We address this gap with a unified benchmark spanning SWE-bench code \citep{jimenez2024swebench}, developer-tool output from Squeez \citep{kovacs2026squeez}, ACL paper chunks \citep{recski2026aclverbatim}, READMEs, Wikipedia markdown \citep{openindex2026wikipedia}, RAGTruth \citep{niu2024ragtruth}, and PsiloQA \citep{rykov2025psiloqa}. The dataset construction uses a shared edit-based labeling step: start from a correct grounded answer, inject a small localized hallucination, recover exact character offsets from the edit, and split by grounding source so the test set uses unseen repositories, papers, or articles. On this benchmark we train two detector families. Our best model is \texttt{LettuceDetect-Qwen-2B}, a fine-tuned Qwen3.5-2B detector \citep{qwen3.5} with a 32{,}768-token maximum sequence length that localizes unsupported spans across code, tool output, structured documents, and natural-language RAG. We also train \texttt{LettuceDetect-mmBERT-base}, a 307M-parameter mmBERT-base encoder \citep{marone2025mmbert}, as a token classification model. Multilingual supervision comes from the 14-language PsiloQA portion of the training set.

Our contributions are:
\begin{itemize}
\setlength{\itemsep}{0pt}
    \item a span-level task formulation for post-generation hallucination detection over structured grounded-generation contexts;
    \item a unified benchmark with 74{,}285 newly constructed examples across code, tool output, and structured documents, plus converted examples from existing natural-language RAG benchmarks;
    \item two detector families that localize unsupported spans across code, tool output, structured documents, and natural-language RAG: \texttt{LettuceDetect-Qwen-2B}, a fine-tuned Qwen3.5-2B detector, and \texttt{LettuceDetect-mmBERT-base}, a 307M-parameter encoder baseline, with results showing that the generative detector substantially outperforms off-the-shelf detectors and zero-shot LLM judges on the code-agent split while remaining competitive on natural-language RAG.
\end{itemize}
Code, data, model checkpoints, prompts, evaluation outputs, and the reviewed code-test arbitration files are released through GitHub and Hugging Face.\footnote{\anonrepo}

\section{Related Work}
\label{sec:related}
\paragraph{Grounded text verification.}
Hallucination detection in grounded generation includes prompt-based judging, self-consistency methods such as SelfCheckGPT \citep{manakul2023selfcheckgpt}, and benchmark-driven detectors trained on datasets such as HaluEval \citep{li2023halueval} and RAGTruth \citep{niu2024ragtruth}. Recent compact detectors, including Luna \citep{belyi2025luna} and LettuceDetect \citep{kovacs2025lettucedetect}, show that long-context encoders can localize unsupported spans at lower cost than LLM judges. We use mmBERT, a ModernBERT-family multilingual encoder trained with annealed language learning \citep{marone2025mmbert}; PsiloQA reports strong span-localization results from fine-tuning mmBERT-base \citep{rykov2025psiloqa}. Other work optimizes related but different objectives: RAG-HAT \citep{song2024raghat} reports response-level F1 on RAGTruth, RL4HS \citep{su2025rl4hs} optimizes span-F1 with reinforcement learning, and PsiloQA \citep{rykov2025psiloqa} and Mu-SHROOM \citep{vazquez2025mushroom} evaluate multilingual span localization. Fine-grained taxonomies have also been proposed, most notably FAVA \citep{mishra2024fava}, but these taxonomies are still mainly designed for natural-language responses grounded in textual documents.

\paragraph{Code hallucination.}
CodeHalu \citep{liu2024codehalu} and CodeMirage \citep{venkatesh2024codemirage} treat hallucinations as defects in generated snippets. Collu-Bench \citep{zhang2024collubench} predicts hallucination during generation from token probabilities and execution feedback. AgentHallu \citep{wang2026agenthallu} attributes failures across agent trajectories. Delulu \citep{erfanian2026delulu} is closest in spirit because it targets code hallucination, but it is an execution-verified fill-in-the-middle benchmark with binary accept/reject labels. In contrast, our task is post-generation, repository-grounded, and span-level.

\section{Task}
\label{sec:task}
Each example consists of a request $q$, context $c$, and answer $a$. The context may contain source code at a specific commit, tool output, or structured document text. The goal is to predict character spans in $a$ that are not supported by $q$ and $c$.

We use three top-level hallucination categories. \textbf{Contradiction} covers wrong logic, values, fields, or conditions in otherwise plausible code. \textbf{Unsupported addition} covers extra behavior or claims not requested or evidenced. \textbf{Fabricated reference} covers invented methods, attributes, keyword arguments, sections, or identifiers. This split follows the common distinction made by prior taxonomies between conflicts with evidence, baseless additions, and invented entities or references \citep{niu2024ragtruth,mishra2024fava}. For code, the first two are mostly judged against the request, while fabricated references are judged against repository and library evidence. The detector does not see generator logits or tool trajectories; it only sees inputs available to an external checker after the answer has been produced.

The top-level labels are paired with 13 subcategories that describe the surface element affected: \texttt{entity}, \texttt{temporal}, \texttt{numerical}, \texttt{value}, \texttt{relational}, \texttt{identifier}, \texttt{section}, \texttt{attribute}, \texttt{claim}, \texttt{behavior}, \texttt{elaboration}, \texttt{subjective}, and \texttt{unspecified}. We choose these subcategories by harmonizing distinctions used in prior taxonomies: RAGTruth's conflict and baseless-information labels \citep{niu2024ragtruth}, FAVA's entity, relation, invented, subjective, and unverifiable labels \citep{mishra2024fava}, code-hallucination categories such as naming, resource, and logic errors \citep{liu2024codehalu}, and the code/tool labels needed by our structured sources. The result keeps the three-way category decision interpretable while preserving the surface-level error types needed for analysis.

\section{Benchmark Construction}
\label{sec:benchmark}
\subsection{Sources}
The benchmark has five newly constructed sources and two incorporated natural-language RAG benchmarks. The programming-oriented sources are \emph{code}, built from SWE-bench \citep{jimenez2024swebench}, and \emph{tool output}, built from Squeez \citep{kovacs2026squeez} containing a query, verbose tool observation, and gold relevant lines. The structured-document sources are \emph{ACL}, built from ACL-Verbatim retrieved paper chunks and questions \citep{recski2026aclverbatim}; \emph{README}, collected from popular GitHub repositories through the GitHub API; and \emph{Wikipedia}, sampled from English \texttt{open-wikipedia-markdown} articles \citep{openindex2026wikipedia}. We also include RAGTruth, a human-annotated word-level hallucination corpus for RAG outputs \citep{niu2024ragtruth}, and PsiloQA, a 14-language span-level hallucination benchmark built from Wikipedia QA \citep{rykov2025psiloqa}, to keep the model tied to established natural-language detection tasks.

All new sources use the same sample abstraction: a prompt containing context and request, an answer, and character-level span annotations over the answer. Context and clean-answer construction differ by source. Code examples use the gold SWE-bench fix; tool-output examples generate a short answer from the Squeez query and relevant lines; ACL examples use the top five retrieved paper chunks as context; README and Wikipedia examples are generated from heading-based markdown chunks. Train, development, and test splits are separated by grounding source.

\begin{table}[t]
\centering
\scriptsize
\setlength{\tabcolsep}{2.5pt}
\begin{tabular}{llrr}
\toprule
\textbf{Source} & \textbf{Modality} & \textbf{Samples} & \textbf{Halluc.} \\
\midrule
Code (SWE-bench) & code & 18{,}524 & 9{,}268 \\
Tool output & tool output & 11{,}365 & 5{,}682 \\
Papers (ACL) & markdown & 5{,}355 & 2{,}677 \\
READMEs & markdown & 13{,}803 & 6{,}900 \\
Wikipedia & markdown & 25{,}238 & 12{,}618 \\
\midrule
\textbf{Total} & & \textbf{74{,}285} & \textbf{37{,}145} \\
\bottomrule
\end{tabular}
\caption{Newly constructed benchmark sources. The complete evaluation also includes converted RAGTruth and PsiloQA examples. New-source splits are train/dev/test = 66{,}368 / 2{,}816 / 5{,}101.}
\label{tab:dataset_stats}
\end{table}

\subsection{Code Source}
SWE-bench provides real GitHub issues, repository metadata, base commits, and gold patches. For each instance we recover the files touched by the gold patch at the base commit and render the gold fix as one coding-assistant answer: a patched function, a changed-line fragment, or a natural-language edit instruction. Clean examples use this answer verbatim. Hallucinated examples contain a small edit to the same answer. We do not include clean and hallucinated versions of the same instance as a pair. This setup differs from snippet-only code hallucination, because the answer is evaluated against a concrete repository state and request.

The three answer renderings are meant to cover the kinds of outputs a developer may ask an assistant for. The \emph{function} rendering gives the largest patched function that fits the length cap, and is closest to a direct code suggestion. The \emph{fragment} rendering gives the changed hunk, preserving the local edit without forcing the model to read a full function. The \emph{edit} rendering gives an instruction such as ``in file X, replace Y with Z'', which is common when agents summarize a patch rather than printing a full diff. Instances whose gold answer is a trivial version bump or too long for the context budget are skipped. This filtering is practical rather than conceptual: if the answer itself consumes the full sequence window, the detector cannot use the repository evidence.

The raw SWE-bench issue text is also rewritten by our pipeline into a short developer request. We keep the technical intent but remove issue-tracker noise, reproduction logs, and long discussions that would make the answer-verification task depend on irrelevant text. This request is included in the detector input. It matters especially for \texttt{unsupported\_addition}, because an extra behavior can be unsupported even when it is syntactically valid and uses real repository symbols.

\subsection{Generation Pipeline}
\label{sec:pipeline_core}
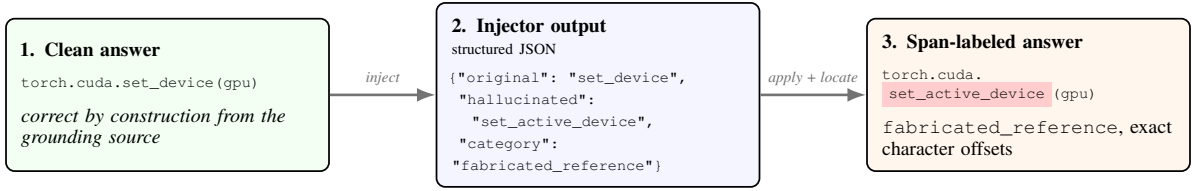
\begin{figure*}[t]
\centering
\definecolor{spanred}{RGB}{255,205,205}
\resizebox{0.98\textwidth}{!}{%
\begin{tikzpicture}[
  font=\footnotesize,
  box/.style={draw, rounded corners, thick, inner sep=7pt, align=left, text width=4.7cm, minimum height=2.45cm},
  arrow/.style={-{Latex[length=3mm]}, very thick, black!55},
]
\node[box, fill=green!5] (clean) {%
  \textbf{1. Clean answer}\\[4pt]
  {\scriptsize\ttfamily torch.cuda.set\_device(gpu)}\\[6pt]
  {\itshape correct by construction from the grounding source}%
};
\node[box, fill=blue!4, right=1.7cm of clean] (json) {%
  \textbf{2. Injector output}\\{\scriptsize structured JSON}\\[3pt]
  {\scriptsize\ttfamily \{"original": "set\_device",}\\
  {\scriptsize\ttfamily ~"hallucinated":}\\
  {\scriptsize\ttfamily ~~~"set\_active\_device",}\\
  {\scriptsize\ttfamily ~"category": "fabricated\_reference"\}}%
};
\node[box, fill=orange!7, right=1.7cm of json] (out) {%
  \textbf{3. Span-labeled answer}\\[4pt]
  {\scriptsize\ttfamily torch.cuda.}\\[-1pt]
  {\scriptsize\ttfamily \colorbox{spanred}{set\_active\_device}(gpu)}\\[6pt]
  {\texttt{fabricated\_reference}, exact character offsets}%
};
\draw[arrow] (clean) -- node[above, font=\scriptsize\itshape]{inject} (json);
\draw[arrow] (json) -- node[above, font=\scriptsize\itshape]{apply + locate} (out);
\end{tikzpicture}%
}
\caption{Edit-based injection yields exact spans. The injector returns each change as an \texttt{original}/\texttt{hallucinated} pair; applying the edit deterministically locates the hallucinated substring in the final answer.}
\label{fig:injection}
\end{figure*}

The construction has a shared labeling step across sources. First, source-specific preparation builds the context, request, and a known-correct grounded answer. Then a source-specific injector proposes localized replacement edits as structured \texttt{original}/\texttt{hallucinated} pairs (Figure~\ref{fig:injection}). Applying these edits gives exact character offsets without diffing mixed natural-language/code answers.

The edit-based design is important for label quality. Instead of diffing two mixed natural-language/code answers, we apply the injector's structured replacement and locate the hallucinated substring directly in the final answer. For the code source, the clean answer is the gold SWE-bench fix and the injector is Gemma 4 31B \citep{google2026gemma4}. For tool output and the markdown sources, Qwen 3.6 35B A3B \citep{qwen2026qwen36} first generates a request or clean answer when the source does not already supply one, and then proposes the hallucination edit. Because the final label comes from the applied replacement, the span can stay narrow: in \texttt{torch.cuda.set\_active\_device(gpu)}, only \texttt{set\_active\_device} is labeled, not the surrounding correct call structure.

\begin{figure*}[t]
\centering
\begin{tcolorbox}[width=0.96\textwidth,colback=gray!4,colframe=black,title={Qualitative benchmark example: contradiction / wrong implementation}]
\small
\textbf{Repository / request.} \texttt{matplotlib\_\_matplotlib-23049}. User request:
\emph{``Can you add support for a \texttt{minor} keyword argument to \texttt{plt.xticks()}?''}

\vspace{3pt}
\textbf{Answer snippet.}
\begin{verbatim}
def xticks(ticks=None, labels=None, *, minor=False, **kwargs):
    ...
    if ticks is None:
        locs = ax.get_xticks(minor=minor)
    else:
        [H] locs = ax.set_xticks(ticks, minor=not minor) [/H]
    ...

def yticks(ticks=None, labels=None, *, minor=False, **kwargs):
    ...
    if ticks is None:
        [H] locs = ax.get_yticks(minor=not minor) [/H]
    ...
\end{verbatim}

\textbf{Type.} Contradiction (\texttt{wrong\_implementation}). The answer is syntactically valid and locally plausible, but reverses the intended behavior of the \texttt{minor} argument.
\end{tcolorbox}
\caption{Example benchmark instance. We use \texttt{[H] ... [/H]} markers for gold unsupported spans.}
\label{fig:qual_example}
\end{figure*}

The injector is prompted to preserve the surrounding answer and to make only localized changes. For code, the target changes fall into two families. Intent errors change a value, field, condition, or side effect while leaving the answer plausible. Structural errors replace a real method, attribute, or keyword with a plausible name that does not exist. The injector returns JSON containing the original substring, the hallucinated substring, and the category. We then apply the replacement deterministically. Attempts are discarded when the original substring is not uniquely found, the edit is a no-op, the hallucinated span covers too much of the answer, or the injected name already occurs in the evidence.

This approach deliberately trades generation yield for cleaner labels. Roughly half of the attempted code injections pass the automatic checks. The most common failures are non-unique originals, overly broad edits, hint words such as ``incorrect'', and fabricated references that are not actually absent from context. These rejections are useful: they remove examples where the model could learn artifacts of the generation process rather than the verification task.

\begin{figure}[t]
\centering
\resizebox{\columnwidth}{!}{%
\begin{tikzpicture}[
  node distance=0.34cm,
  box/.style={draw, rounded corners, thick, align=left, font=\scriptsize, inner sep=4.5pt, text width=6.1cm},
  arrow/.style={-{Latex[length=2mm]}, thick, black!55},
]
\node[box, fill=blue!6] (ref) {\textbf{Answer references}\\
\texttt{self.save\_checkpoint}, \texttt{deprecated}, \texttt{torch.cuda.set\_device}};
\node[box, fill=green!8, below=of ref] (patch) {\textbf{Repository evidence}\\
changed functions and full changed files at the base commit};
\node[box, fill=green!8, below=of patch] (imports) {\textbf{Import graph evidence}\\
modules imported by the answer or by changed files};
\node[box, fill=orange!12, below=of imports] (libs) {\textbf{Third-party evidence}\\
compact library signatures for external APIs};
\node[box, fill=gray!10, below=of libs] (ctx) {\textbf{Verifier context}\\
original prompt plus referenced definitions / library signatures};
\draw[arrow] (ref) -- (patch);
\draw[arrow] (patch) -- (imports);
\draw[arrow] (imports) -- (libs);
\draw[arrow] (libs) -- (ctx);
\end{tikzpicture}%
}
\caption{Reference grounding for code examples. Correct references missing from the truncated source context are resolved at the base commit and appended to the verifier context; third-party APIs are grounded with compact signatures.}
\label{fig:grounding}
\end{figure}
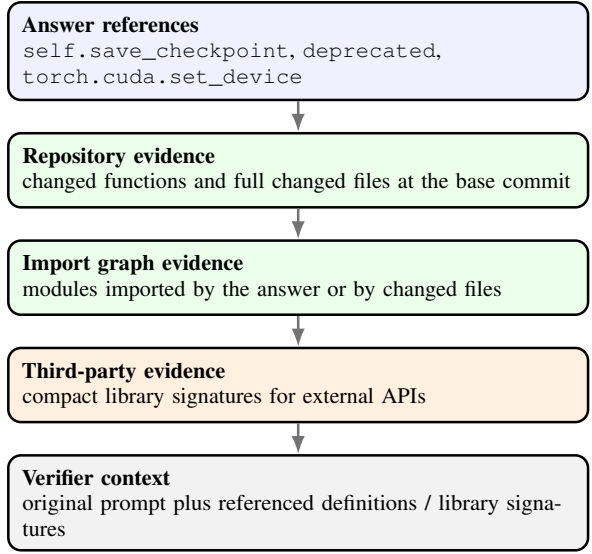

\paragraph{Reference grounding.}
Clean answers often refer to sibling methods, imported helpers, or third-party APIs not present in a truncated source context. If these are not added, a verifier cannot distinguish a correct but unseen reference from a fabricated one. We therefore append referenced definitions from the modified files, imported repository modules, and modules imported by the changed files, all resolved at the historical base commit rather than by current-code search. For third-party APIs, which are not in the repository at all, we retrieve real signatures from Context7\footnote{\url{https://context7.com}}, a library-documentation index: we parse the external libraries imported by the answer and query Context7 for each imported symbol's signature and usage, then append the returned snippets (Figure~\ref{fig:grounding}). This gives the verifier genuine evidence for external calls, so a real but unfamiliar library API is not penalized as a fabrication.

\paragraph{Other sources.}
The non-code sources use the same edit-application framework but different prompts. Tool-output injections misreport identifiers, line references, values, or claims from the observation. ACL injections use paper-specific numerical, entity, relational, methodological, and citation-like edits detectable from the retrieved excerpts. README and Wikipedia injections use a generic markdown prompt covering numerical, temporal, entity, relational, fabricated-reference, and unsupported-claim edits.

\subsection{Test-Set Verification}
\label{sec:test_verification}
For quality assurance, we reviewed every code-source test sample before release: 2{,}038 examples were reviewed and 2{,}015 retained. Model-assisted triage first flagged span validity, category, boundary, and plausibility issues\footnote{We used Claude Sonnet 4.6 to flag candidate issues; final decisions were made by the authors during arbitration.}; flagged cases were then re-judged blindly and finally arbitrated by the authors against the true pre-fix repository evidence rather than the injector's intended edit. For the 44 disputed hallucinations, arbitration upheld 41 as genuine and dropped 3 that matched original code. The review tightened 235 boundaries, dropped 23 invalid spans, corrected 2 categories, reclassified 5 examples as clean, and removed 23 examples for question--answer mismatch, no-op edits, or incoherent rendering. No rebalancing was applied after review; the released code test set is 50.3\% hallucinated. Per-sample verdicts, contested cases, resolutions, and the rubric are released with the dataset.

\section{Benchmark Characteristics}
\label{sec:characteristics}
The code source contains 18{,}524 examples from 53 repositories, split by SWE-bench repository into 16{,}319 train / 190 development / 2{,}015 test examples over 35 / 6 / 12 disjoint repositories (50.0\% hallucinated overall). Average answers are about 1.8k characters, and hallucinated code examples contain just under two labeled spans on average. Answer formats are uneven---function (8{,}550), fragment (4{,}893), and edit-style (5{,}081) renderings---as are error types over hallucinated examples: contradiction (4{,}719), fabricated reference (2{,}275), and unsupported addition (2{,}274). 

The examples vary in what evidence is needed. Some hallucinations are local: a wrong config key is contradicted by the nearby code. Others require cross-file evidence, for example a method called on \texttt{self} whose definition lives in an imported mixin. Third-party API fabrications require library signatures rather than repository code. Without reference grounding, correct answers would contain many unsupported-looking names simply because the context was truncated.

For the tool-output dataset the context is often an observation rather than a source file: a test failure, a grep result, a package-manager response, or a git command. Here hallucinations are usually value and relation errors: reporting the wrong file, count, version, status, or failing test. The markdown sources broaden the data beyond code while preserving structure. ACL papers, READMEs, and Wikipedia pages contain headings, tables, citations, and lists, so the detector sees contexts that are not plain paragraphs but are still text-grounded.

\section{Experiments}
\label{sec:experiments}
\subsection{Models and Metrics}
Our best model, \texttt{LettuceDetect-Qwen-2B}, is a Qwen3.5-2B generative detector \citep{qwen3.5} fine-tuned to return JSON spans with category and subcategory. We fine-tune and evaluate it with a 32{,}768-token maximum sequence length, so each prediction can include the user request, retrieved documents or repository evidence, tool output, and the answer to check. It is trained on the unified training set using the same data-agnostic prompt across sources. We compare it with \texttt{LettuceDetect-mmBERT-base}, a 307M-parameter mmBERT-base encoder \citep{marone2025mmbert} trained on the same task, an LFM2.5-8B-A1B generative sibling \citep{liquidai2026lfm25} fine-tuned with the same prompt, LettuceDetect-large \citep{kovacs2025lettucedetect}, answer-level faithfulness detectors, and two large zero-shot LLM judges: Nemotron-3-Ultra-550B \citep{nvidia2026nemotron3ultra} and gpt-oss-120b \citep{openai2025gptoss}.

We report character-overlap span precision, recall, and F1; example-F1, where an answer is flagged if at least one span is predicted; and mean IoU between predicted and gold span mass. For typed detection, a predicted span only receives credit when its category or subcategory also matches.

\paragraph{Training details.}
The generative detector is trained with supervised fine-tuning on the merged structured and natural-language training split. The 74{,}285 examples in Table~\ref{tab:dataset_stats} are newly constructed; after adding converted RAGTruth and PsiloQA examples, the full split contains 145{,}250 train, 6{,}171 validation, and 10{,}698 test examples. The training and evaluation splits are released on Hugging Face, and the construction/evaluation code is released on GitHub.\footnote{\anonrepo} We fine-tune Qwen3.5-2B with LoRA rank 32, $\alpha=64$, dropout 0, bf16 weights, learning rate $2{\times}10^{-4}$, linear schedule with 3\% warmup, weight decay 0.01, two epochs, effective batch size 32, and a 32{,}768-token maximum sequence length. The multilingual portion of the training data comes from PsiloQA, while the long context budget is mainly used by code, tool-output, and structured-document examples. The prompt is the same across sources: it defines a hallucinated span, lists the taxonomy, then provides the user request, context, and answer to verify. Returned strings are matched back into the answer to recover character offsets.

\texttt{LettuceDetect-mmBERT-base} follows the LettuceDetect token-classification architecture with \texttt{jhu-clsp/mmBERT-base} as the backbone: context and request tokens are masked in the loss, and answer tokens receive supported/unsupported labels. We train it on the same unified split with an 8{,}192-token maximum sequence length, batch size 8, gradient accumulation 4, learning rate $10^{-5}$, and three epochs, selecting the best checkpoint by development hallucinated-token F1. The encoder is cheaper at inference and decodes contiguous positive answer tokens into character spans, but it only predicts binary spans in its base form. We also evaluate a typed encoder cascade in which spans from the frozen binary model are classified by a label-conditioned head that scores each span against category descriptions.

\subsection{Main Results}
Table~\ref{tab:main_results} gives per-source results for the 2B generative detector. The model performs best on README and Wikipedia, where grounding resembles factual document QA, and remains strong on the harder coding-agent sources. Code-agent is the most difficult source: the context is long, the answer often contains new code, and some errors are intent mistakes rather than simple factual contradictions.

\begin{table*}[t]
\centering
\small
\setlength{\tabcolsep}{5pt}
\begin{tabular}{lrrrrrrrr}
\toprule
 & & \multicolumn{4}{c}{\textbf{LettuceDetect-Qwen-2B}} & \textbf{LD-mmBERT} & \textbf{LFM-8B} & \textbf{gpt-oss} \\
\cmidrule(lr){3-6}\cmidrule(lr){7-7}\cmidrule(lr){8-8}\cmidrule(lr){9-9}
\textbf{Source} & \textbf{n} & \textbf{P} & \textbf{R} & \textbf{F1} & \textbf{Ex.-F1} & \textbf{F1} & \textbf{F1} & \textbf{F1} \\
\midrule
ALL & 10{,}698 & 0.677 & 0.701 & \textbf{0.689} & 0.921 & 0.642 & 0.650 & -- \\
Code-agent & 2{,}015 & 0.596 & 0.609 & \textbf{0.602} & 0.835 & 0.508 & 0.507 & 0.177 \\
Tool output & 617 & 0.737 & 0.702 & \textbf{0.719} & 0.907 & 0.588 & 0.692 & 0.331 \\
ACL & 440 & 0.752 & 0.745 & \textbf{0.749} & 0.942 & 0.579 & 0.670 & 0.602 \\
README & 641 & 0.889 & 0.844 & \textbf{0.866} & 0.984 & 0.751 & 0.804 & 0.666 \\
Wikipedia & 1{,}388 & 0.836 & 0.800 & \textbf{0.817} & 0.974 & 0.708 & 0.789 & 0.658 \\
PsiloQA & 2{,}897 & 0.702 & 0.765 & \textbf{0.732} & 0.966 & 0.714 & 0.714 & -- \\
RAGTruth & 2{,}700 & 0.601 & 0.548 & \textbf{0.574} & 0.818 & 0.528 & 0.557 & -- \\
\bottomrule
\end{tabular}
\caption{Per-source results. P/R/F1/Ex.-F1 are character-overlap span metrics for LettuceDetect-Qwen-2B; LD-mmBERT, LFM-8B, and gpt-oss report span-F1. gpt-oss was run only on the five newly constructed sources.}
\label{tab:main_results}
\end{table*}

The generative detector outperforms LettuceDetect-mmBERT-base on every source: $0.689$ vs.\ $0.642$ span-F1 overall, and $0.602$ vs.\ $0.508$ on code-agent. On code-agent answers it is also well above LettuceDetect-large and the strongest zero-shot LLM judge we evaluated; on natural-language RAG benchmarks it is close to specialized systems.

\subsection{Comparison with Existing RAG Benchmarks}
Table~\ref{tab:prose_comparison} compares against published natural-language RAG results. On RAGTruth, our model is highly competitive with specialized systems while also covering code and tool output. On PsiloQA, it sets the best reported English IoU ($0.724$), above the fine-tuned mmBERT-base model ($0.707$) of the benchmark authors and far above their 32B few-shot judge ($0.400$). Across all 14 PsiloQA languages, it reaches $0.689$ mean IoU, compared with $0.623$ for the PsiloQA mmBERT-base model and $0.383$ for the Qwen2.5-32B judge reported by \citet{rykov2025psiloqa}. These are the strongest reported results among the published PsiloQA systems we compare against, while using the same detector trained for code, tool output, and structured documents.

\begin{table}[t]
\centering
\scriptsize
\setlength{\tabcolsep}{3pt}
\begin{tabularx}{\columnwidth}{@{}p{0.31\columnwidth}Xr@{}}
\toprule
\textbf{Benchmark} & \textbf{Method} & \textbf{Score} \\
\midrule
RAGTruth Ex.-F1 & RAG-HAT 8B \citep{song2024raghat} & 83.9 \\
RAGTruth Ex.-F1 & LettuceDetect-Qwen-2B & 81.8 \\
RAGTruth Ex.-F1 & LettuceDetect-large \citep{kovacs2025lettucedetect} & 79.2 \\
RAGTruth Ex.-F1 & Fine-tuned Llama-2-13B \citep{niu2024ragtruth} & 78.7 \\
RAGTruth Ex.-F1 & GPT-4 \citep{niu2024ragtruth} & 63.4 \\
\midrule
PsiloQA EN IoU & LettuceDetect-Qwen-2B & 0.724 \\
PsiloQA EN IoU & mmBERT-base (PsiloQA) \citep{rykov2025psiloqa} & 0.707 \\
PsiloQA EN IoU & Qwen2.5-32B judge \citep{rykov2025psiloqa} & 0.400 \\
\midrule
PsiloQA 14-lang IoU & LettuceDetect-Qwen-2B & 0.689 \\
PsiloQA 14-lang IoU & mmBERT-base (PsiloQA) \citep{rykov2025psiloqa} & 0.623 \\
PsiloQA 14-lang IoU & Qwen2.5-32B judge \citep{rykov2025psiloqa} & 0.383 \\
\bottomrule
\end{tabularx}
\caption{Comparison with established natural-language RAG benchmarks. Rows combine published numbers with our evaluation, so they should be read as external reference points rather than a single shared leaderboard.}
\label{tab:prose_comparison}
\end{table}

\subsection{Code and Tool Evidence Evaluation}
Table~\ref{tab:code_baselines} shows results on the code-agent test set. LettuceDetect-large, trained for natural-language RAG, reaches only $0.17$ span-F1. Large zero-shot judges find some true error regions but over-predict heavily: the task-aware Nemotron prompt improves over a generic prompt but still reaches only $0.22$ span-F1. Answer-level faithfulness systems, including HHEM-2.1-Open \citep{li2024hhem}, Lynx-8B \citep{ravi2024lynx}, Granite-Guardian-4.1-8B \citep{padhi2025graniteguardian}, and MiniCheck-7B \citep{tang2024minicheck}, show the same tendency at answer level: high recall but much lower precision on the hallucinated class. In most cases these judges flag correct newly written patch code as unsupported instead of checking whether the edit follows from the request and repository evidence.

Table~\ref{tab:main_results} shows results on other sources. Under the same generic prompt, gpt-oss-120b reaches reasonable span-F1 on document-like data (ACL, README, Wikipedia), but performs poorly on tool output and code-agent answers \citep{openai2025gptoss}.

\begin{table}[t]
\centering
\scriptsize
\setlength{\tabcolsep}{2.2pt}
\begin{tabular}{llrrrr}
\toprule
\textbf{Detector (code-agent)} & \textbf{Eval} & \textbf{P} & \textbf{R} & \textbf{F1} & \textbf{Ex.-F1} \\
\midrule
LettuceDetect-large (v1) & span & 0.112 & 0.373 & 0.172 & 0.684 \\
Nemotron-3-Ultra, generic & span & 0.108 & 0.665 & 0.186 & 0.655 \\
Nemotron-3-Ultra, task-aware & span & 0.132 & 0.589 & 0.216 & 0.700 \\
gpt-oss-120b, task-aware & span & 0.135 & 0.501 & 0.212 & 0.691 \\
\midrule
HHEM-2.1-Open & answer & 0.50 & 0.86 & 0.63 & -- \\
Lynx-8B & answer & 0.52 & 0.73 & 0.61 & -- \\
Granite-Guardian-4.1-8B & answer & 0.53 & 0.90 & 0.66 & -- \\
MiniCheck-7B & answer & 0.50 & 1.00 & 0.67 & -- \\
\midrule
LettuceDetect-mmBERT-base & span & 0.619 & 0.430 & 0.508 & 0.770 \\
LFM2.5-8B-A1B (ours, SFT) & span & 0.531 & 0.485 & 0.507 & 0.811 \\
\textbf{LettuceDetect-Qwen-2B} & span & \textbf{0.596} & \textbf{0.609} & \textbf{0.602} & \textbf{0.835} \\
\bottomrule
\end{tabular}
\caption{Baselines on the code-agent test set. Span systems report character-overlap span P/R/F1 plus example-F1; answer systems report hallucinated-class example P/R/F1.}
\label{tab:code_baselines}
\end{table}

\subsection{Typed Detection}
The generative detector emits a category and subcategory with every span. On the full unified test set it reaches category-gated span-F1 $0.585$ and subcategory-gated span-F1 $0.468$, compared with binary span-F1 $0.689$. Subcategory prediction is harder, especially for natural-language examples where \texttt{claim}, \texttt{elaboration}, \texttt{value}, and \texttt{relational} can overlap. As a cheaper alternative, we also evaluate a typed encoder cascade: given gold spans, the label-conditioned head reaches $0.82$ category and $0.64$ subcategory accuracy, but end-to-end it trails the generative model (category-gated span-F1 $0.461$ vs.\ $0.585$, subcategory-gated $0.315$ vs.\ $0.468$ overall).

\subsection{Error Analysis}
The hardest remaining code examples are request-grounded intent errors and broad generated edits. A wrong field name or condition may use real repository symbols and look structurally valid, while only one small part of a larger generated block is unsupported. Tool-output examples show the same pattern when an otherwise fluent answer copies one version, count, filename, or status incorrectly.

Zero-shot judges make similar errors. With a generic prompt, Nemotron-3-Ultra often marks clean patch code as fabricated when it should verify newly written lines against the request and repository evidence. A task-aware prompt improves precision from 0.11 to 0.13 but does not solve the problem, and gpt-oss-120b behaves similarly. In a small reasoning-judge diagnostic, correct patch lines were treated as unsupported, truncated context as evidence for fabricated references, and edit-style answers produced non-verbatim spans. These results show that strong zero-shot judges do not replace task-specific training here.

\section{Conclusion}
\label{sec:conclusion}
We introduced a span-level hallucination detection benchmark across code, tool output, structured documents, and natural-language RAG. A 2B generative detector reaches $0.689$ span-F1 overall and $0.602$ on code-agent answers, well above the off-the-shelf detectors and zero-shot LLM judges we evaluated. It also reaches the best reported English PsiloQA IoU ($0.724$) and $81.8$ RAGTruth example-F1, close to RAG-HAT's $83.9$.

\section{Limitations}
\label{sec:limitations}
Most labels come from synthetic injection. The reviewed code test set gives us confidence in that split, but train/development labels and non-code test labels are generated labels guarded by automated checks, and the code review is model-assisted rather than independently annotated by multiple human annotators. The benchmark covers final-answer verification, not full agent trajectories, and should not be read as measuring real-world hallucination prevalence.

\bibliography{references}

\appendix

\section{Prompt Templates}
\label{app:prompts}

We release the exact prompt code with the dataset. This appendix gives the main templates used for detector training and synthetic data construction. Variable slots are shown as \texttt{\{name\}}; lines beginning with \texttt{\#\#} are annotations in the figure, not part of the prompt.

\begin{figure*}[htbp]
\begin{lstlisting}[style=prompt]
You are an expert annotator who identifies hallucinated spans in a generated answer
with respect to a given context (the only trusted evidence). A hallucinated span is a
substring of the answer that is not supported by the context. Spans consistent with the
context are not hallucinations.

Quote each hallucinated span verbatim from the answer and classify it into exactly one
category and one subcategory.

Categories:
- contradiction: conflicts with the context (a wrong value, number, date, name, or relationship)
- fabricated_reference: an entity, name, identifier, or section that is absent from the context
- unsupported_addition: a claim, detail, or behavior the context never states

Subcategories:
entity, temporal, numerical, value, relational, identifier, section, attribute,
claim, behavior, elaboration, subjective, unspecified

Reply with ONLY a JSON object:
{"hallucinated_spans": [{"text": "...", "category": "...", "subcategory": "..."}]}.
If nothing is unsupported, reply {"hallucinated_spans": []}.

User message:
{request_and_context}

Answer to verify:
{answer}
\end{lstlisting}
\caption{Detector prompt used for generative detector training and zero-shot LLM-judge evaluation. For code-agent rows in the generative detector training set, the same prompt also requests a short explanation field for each span; clean and hallucinated rows from that source both use the explanation variant, so the prompt choice does not leak the label.}
\label{fig:prompt-detector}
\end{figure*}

\begin{figure*}[htbp]
\begin{lstlisting}[style=prompt]
## Question generation for README and Wikipedia sources
You generate a single, natural information-seeking question that can be answered
from a given document.

Document:
{document_chunk}

Generate one {question_type} question ({question_type_definition}) that the
document answers.

Rules:
1. Return ONLY the question, nothing else.
2. Use neutral, self-contained phrasing.
3. Keep it short and natural, like a query typed into a search engine.
4. It must be answerable from the document above.

## Clean-answer generation for tool-output, ACL, README, and Wikipedia sources
You are a helpful assistant answering a user's question using ONLY the provided
evidence. Write a correct, natural answer grounded strictly in that evidence.

Your answer MUST:
- Be accurate and fully supported by the evidence -- invent nothing.
- Reference concrete details from the evidence where relevant.
- Be concise.

Your answer must NOT:
- Add claims, identifiers, or values not present in the evidence.
- Include filler.
\end{lstlisting}
\caption{Question and clean-answer generation prompts. README and Wikipedia examples first generate a self-contained question from a markdown chunk; tool-output, ACL, README, and Wikipedia examples then generate a clean answer grounded only in the supplied evidence.}
\label{fig:prompt-generation}
\end{figure*}

\begin{figure*}[htbp]
\begin{lstlisting}[style=prompt]
You are a hallucination injector for building a hallucination detection dataset.

You are given a CORRECT answer and the CONTEXT it is grounded in. Return ONLY a
small set of localized replacement edits that turn the answer into one containing
a specific kind of hallucination. The pipeline applies your edits; outside them
the answer must stay identical.

Target hallucination:
- Category: {category}
- Subtype: {subcategory}

CRITICAL grounding rule:
- The injected error MUST be detectable by comparing the answer against the
  provided context alone.

Rules:
- Make 1-2 distinct edits targeting the subtype above.
- Each replacement span must be as small as possible.
- Total changed text must be less than 30% of the answer.
- Changes must be plausible and subtle.
- Each "original" must be an exact substring of the answer, appearing exactly once.

Respond in JSON:
{"changes": [{"original": "...", "hallucinated": "...", "explanation": "..."}]}
\end{lstlisting}
\caption{Generic injection prompt. The model proposes structured replacement edits; the pipeline applies them deterministically and recovers exact character offsets. Code uses two source-specific variants, one for wrong implementations and unrequested changes and one for fabricated methods, attributes, and keyword arguments. Both keep the same \texttt{original}/\texttt{hallucinated} output format.}
\label{fig:prompt-injection}
\end{figure*}

\end{document}